\documentclass[letterpaper, 10 pt, conference]{ieeeconf}  % Comment this line out if you need a4paper

\IEEEoverridecommandlockouts                              % This command is only needed if 
\usepackage{graphics} % for pdf, bitmapped graphics files
\usepackage{epsfig} % for postscript graphics files
\usepackage{mathptmx} % assumes new font selection scheme installed
\usepackage{times} % assumes new font selection scheme installed
\usepackage{amsmath} % assumes amsmath package installed
\usepackage{amssymb}  % assumes amsmath package installed
\usepackage{color}
\usepackage{booktabs}
\title{\LARGE \bf
GestAdapt: Workspace-Conditioned Co-Speech Gesture Generation for Humanoid Robots
}

\newif\ifanonymous
\ifanonymous
    \author{}
\else
    \author{%
        Bosong Ding$^{1,*}$,
        Xianglin Zhang$^{2}$,
        Miao Xin$^{3}$,
        Murat Kirtay$^{1}$,
        and Giacomo Spigler$^{1}$%
        \thanks{$^{*}$Corresponding author.}%
        \thanks{$^{1}$Department of Intelligent Systems, Tilburg University, The Netherlands.}%
        \thanks{$^{2}$TU-Dresden, Dresden, Germany.}
        \thanks{$^{3}$School of Artificial Intelligence, China University of Mining and Technology-Beijing (CUMTB), Beijing, China.}%
    }
\fi

\begin{document}

\maketitle
\thispagestyle{empty}
\pagestyle{empty}

%%%%%%%%%%%%%%%%%%%%%%%%%%%%%%%%%%%%%%%%%%%%%%%%%%%%%%%%%%%%%%%%%%%%%%%%%%%%%%%%

\begin{abstract}
Co-speech gestures for robots must adapt not only to speech and embodiment, but also to the workspace available for performing the motion. Since the same speech can be accompanied by different gestures, a robot can respond to workspace constraints, e.g., gestures for speech next to a wall. In these scenarios, the robot should gesture in a suitable motion rather than simply correcting an unconstrained one. To achieve this goal, we present GestAdapt, a workspace-conditioned framework that conditions co-speech gesture generation on a prescribed wrist workspace. The GestAdapt framework learns from six complementary co-speech corpora through a shared motion representation and supports retargeting to different robot embodiments. Quantitative evaluation shows that generated motions remain close to the real-motion distribution while respecting the workspace. In a user study, gestures generated under modified workspace constraints receive a mean quality score of 3.24/5, above our no-workspace variant (2.43/5) and below the reference motions (3.68/5). In a real robot evaluation, all compared motions are retargeted to the Reachy2 humanoid robot under identical workspace constraints. Motions generated with our framework rank first in 69.7\% of comparisons, higher than our no-workspace variant baseline and retargeted ground-truth motions constrained afterward. Overall, the results support adapting gestures to the available workspace during generation, rather than modifying unconstrained trajectories afterward to satisfy workspace constraints, potentially compromising gesture naturalness.
\end{abstract}

%%%%%%%%%%%%%%%%%%%%%%%%%%%%%%%%%%%%%%%%%%%%%%%%%%%%%%%%%%%%%%%%%%%%%%%%%%%%%%%%
\section{INTRODUCTION}

Co-speech gestures are body movements that naturally accompany speech, and co-speech gesture generation aims to synthesize such movements from speech \cite{nyatsanga2023review}. However, the space available for gesturing can vary substantially across environments. For example, a person may express the same content while standing in an open room, sitting across a table, or standing close to a wall. Although the speech remains unchanged, the surrounding space constrains the range of feasible gestures. Humanoid robots face similar limitations. A gesture-generation model without spatial conditioning may produce broad arm movements that are inappropriate for the robot's surroundings or limited joint ranges. Consequently, a speech-conditioned model that does not account for constraints may generate motions that reduce motion quality, violate the robot's physical limits, or even risk damaging the hardware. In such settings, a gesture that appears natural in the training data may therefore be difficult or unsafe for the robot to execute.

\begin{figure}[h]
  \centering
  \includegraphics[width=\linewidth]{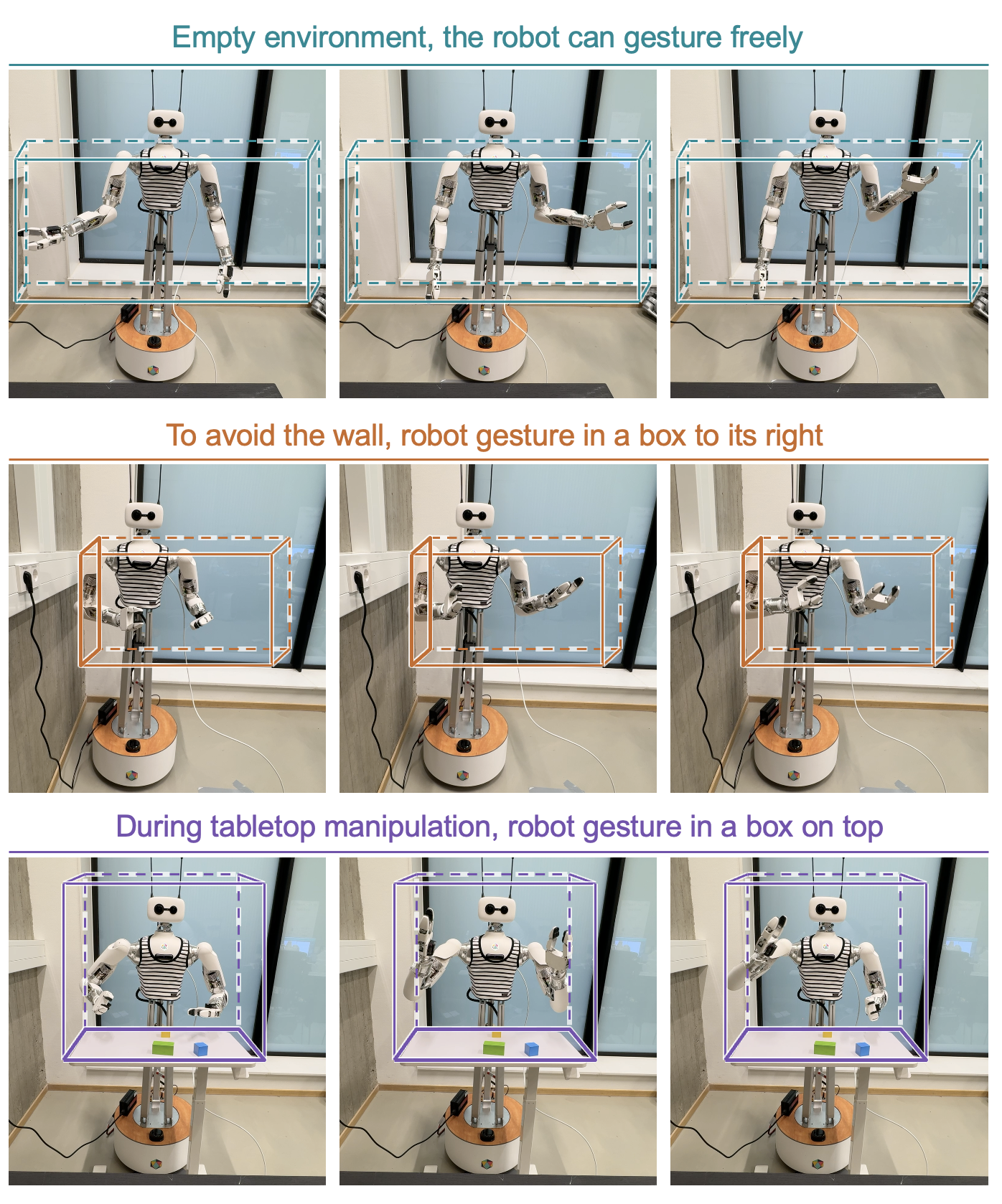}
\caption{
Examples of workspace-adaptive co-speech gestures by a Reachy2 humanoid robot.
The chosen gesture space changes across three settings:
(a) free space, (b) a nearby wall, and (c) a tabletop workspace.
The robot adapts its gesture motion to remain within the specified workspace.
}
\label{fig:workspace_examples}
\end{figure}

Most co-speech gesture models focus on naturalness, synchronization, semantics, style, and diversity \cite{yoon2020trimodal,alexanderson2020style,liu2022beat,zhu2023diffgesture,liu2024emage}. A smaller body of work has brought co-speech gesture generation onto humanoid robots \cite{yoon2019robots,liang2026physdrift,wang2026robogesture,ding2026cospeech}. These works additionally require that generated motion remain executable under the robot's embodiment and kinematic constraints and emphasize motion transfer, embodiment-aware generation, kinematic feasibility, and safe execution. Despite the growing number of studies, generating gestures that respect the limitations of the surrounding environment has been identified as an important open challenge in co-speech gesture generation \cite{nyatsanga2023review}. More recent approaches have begun to address this problem for human or virtual-character motion: InteracTalker combines co-speech motion with object-aware interaction \cite{rajan2026interactalker}, while Puppeteer directly conditions gesture generation on posture and surrounding object geometry \cite{adeli2026puppeteer}. This leaves a vital problem at the intersection of the two directions: generating co-speech motion according to the physical space available to a robot while considering its embodiment and execution constraints.

Co-speech gesture generation is inherently one-to-many: the same speech can be accompanied by different hand configurations, movement directions, and amplitudes while remaining natural and communicative \cite{li2021audio2gestures,zhu2023diffgesture}. Existing methods already exploit this variability by conditioning on speaker, motion style, or partially specified motion \cite{yoon2020trimodal,alexanderson2020style,ghorbani2023zeroeggs,liu2024emage}. For a robot, physical cues can serve as an additional conditioning signal. Together, these allow the model to generate a different but plausible gesture that naturally fits the available space, without needing to compress or correct the motion, as shown in Fig. \ref{fig:workspace_examples}.

To this end, we present GestAdapt, a workspace-conditioned framework for robot co-speech gesture generation. We combine six co-speech corpora in a unified upper-body and hand representation. Spatial supervision comes from the wrist-motion envelopes without requiring scene annotations. A diffusion Transformer generates future motion conditioned on speech, motion history, and a specified workspace box. Learned conditioning allows the gesture itself to change with the available space, while sampling-time analytic guidance reduces residual wrist violations. The generated motion is then transferred to the target robot through embodiment-specific retargeting and executed with constrained workspace limits and hard physical constraints.

We evaluate GestAdapt through quantitative motion metrics, human evaluation under changed workspace conditions, and physical robot experiments. The results show that the model can adapt gesture motion to new spatial constraints while preserving plausible and smooth motion. In the robot evaluation, where all compared motions undergo the same retargeting and constrained execution pipeline, participants prefer workspace-conditioned generation over the same generator trained without workspace conditioning and directly retargeted human motion. Together, these results support treating target space as part of gesture generation itself, rather than only as a constraint imposed afterward.

\section{RELATED WORK}

\subsection{Co-Speech Gesture Generation}

Learning-based co-speech gesture generation has progressed from predicting motion from speech acoustics to incorporating linguistic content, speaker identity, and multimodal speech representations \cite{ginosar2019gestures,kucherenko2020gesticulator,yoon2020trimodal}. Recent work has further improved holistic body-hand coordination, style control, and motion diversity \cite{yi2023talkshow,alexanderson2020style,li2021audio2gestures,zhu2023diffgesture,liu2024emage}.

At the same time, a number of co-speech motion corpora have been
introduced, including BEAT \cite{liu2022beat}, SHOW \cite{yi2023talkshow}, Streamer \cite{yang2025gesturehydra}, TED \cite{liu2022learning}, Trinity \cite{ferstl2018investigating}, and ZEGGS \cite{ghorbani2023zeroeggs}. These datasets were collected independently with different acquisition setups, motion representations, and gesture distributions; as a result, most prior gesture models are developed within a particular corpus and its associated representation.

Unlike these studies, our GestAdapt framework unifies multiple corpora in a shared motion space, using their complementary motion and spatial variation to support general workspace-conditioned gesture generation.

\subsection{Gesture Generation for Robots}

Early learning-based systems demonstrated speech-driven gesture generation on humanoid robots \cite{yoon2019robots,yu2020srg3}. Transferring human motion to a robot requires accounting for differences in limb proportions, joint configurations, and motion limits, often through retargeting and constrained optimization \cite{dariush2008retargeting}. More recent work incorporates embodiment requirements more directly. PhysDrift~\cite{liang2026physdrift}, for example, learns robot-native joint trajectories from speech using embodiment-aware motion curation and regularization. RoboGesture~\cite{wang2026robogesture} combines robot-centric gesture generation with semantic-acoustic alignment and an MPC-based collision-avoidance filter for execution. These methods address important challenges in embodiment transfer, speech alignment, and physical execution. We study a different but complementary question: \textbf{how should the gesture itself change when the space available to the robot changes?} Our model conditions generation on the target workspace before embodiment-specific retargeting and constrained execution.

\subsection{Physical and Spatial Constraints in Motion Generation}

Scene-conditioned human motion models incorporate environmental geometry for locomotion and human-object interaction \cite{hassan2021samp,zhao2023dimos}, while robot motion planning routinely considers kinematic and collision constraints \cite{kuffner2001motion,ratliff2009chomp}. Spatial context has also begun to enter the co-speech gesture generation literature. InteracTalker integrates speech-driven gestures with prompt-based object interactions \cite{rajan2026interactalker}, and Puppeteer conditions gesture synthesis on posture and surrounding object geometry \cite{adeli2026puppeteer}. Our setting differs regarding its supervision and deployment requirements. Instead of learning from scene-annotated gesture recordings, we derive weak spatial conditions from wrist-motion envelopes across existing speech-motion corpora. We use an explicit workspace box to control gesture extent and evaluate the generated motions after robot retargeting under shared execution constraints. This connects spatially conditioned gesture generation with embodiment-specific robot gesture execution without requiring a full scene representation.

\section{Unified Dataset}

\begin{figure*}[h]
  \centering
  \includegraphics[width=\linewidth]
  {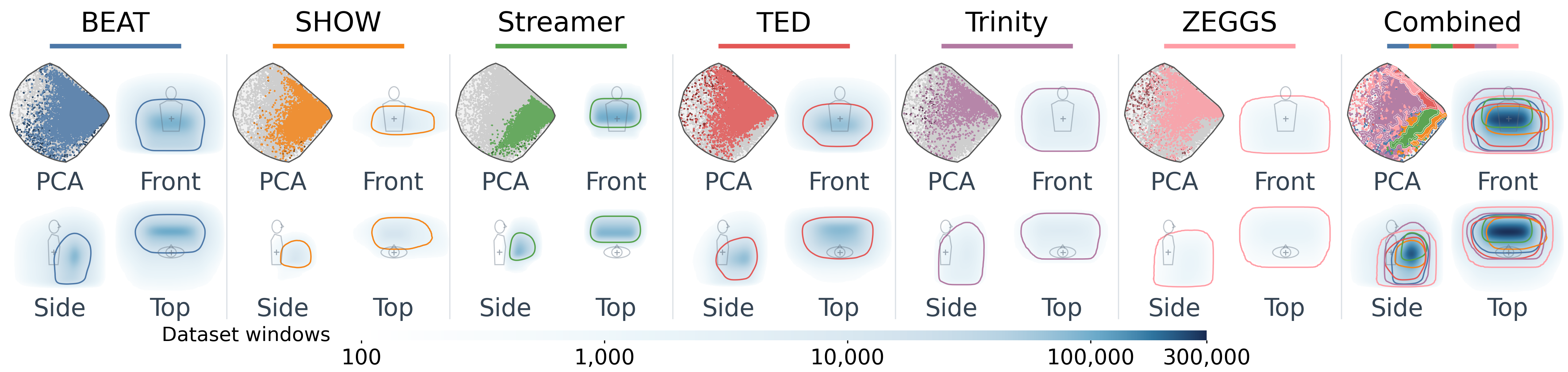}
  \caption{
  Complementary motion and spatial support across the six co-speech
  corpora. The Principal Component Analysis (PCA) views show the motion distributions, while the
  front, side, and top views visualize the corresponding wrist-motion
  envelopes. Heatmap intensity indicates the number of envelopes
  covering each location, and colored contours show the spatial support
  of individual corpora.
  }
  \label{fig:sixdataset}
\end{figure*}

\subsection{Combining Multiple Co-Speech Corpora}
Learning spatially adaptive gestures ideally requires the same speech to be performed under different physical constraints. Existing co-speech datasets do not provide such paired recordings or explicit free-space annotations. Therefore, we combine six datasets with aligned speech and motion: BEAT \cite{liu2022beat}, SHOW \cite{yi2023talkshow}, Streamer \cite{yang2025gesturehydra}, TED-Expressive \cite{liu2022learning}, Trinity \cite{ferstl2018investigating}, and ZEGGS \cite{ghorbani2023zeroeggs}.

These datasets differ in recording settings and gesture distributions, providing complementary variation in posture, gesture scale, and upper-body workspace. As shown in Fig.~\ref{fig:sixdataset}, their motion distributions overlap, while their kinematic and spatial support remain distinct. Combining them expands both the motion and workspace coverage available for training. Since their native motion representations are incompatible, we first map them into a shared geometric representation.

\subsection{Robot-Centered Unified Motion Representation}
\begin{figure}[h]
  \centering
  \includegraphics[width=\linewidth]{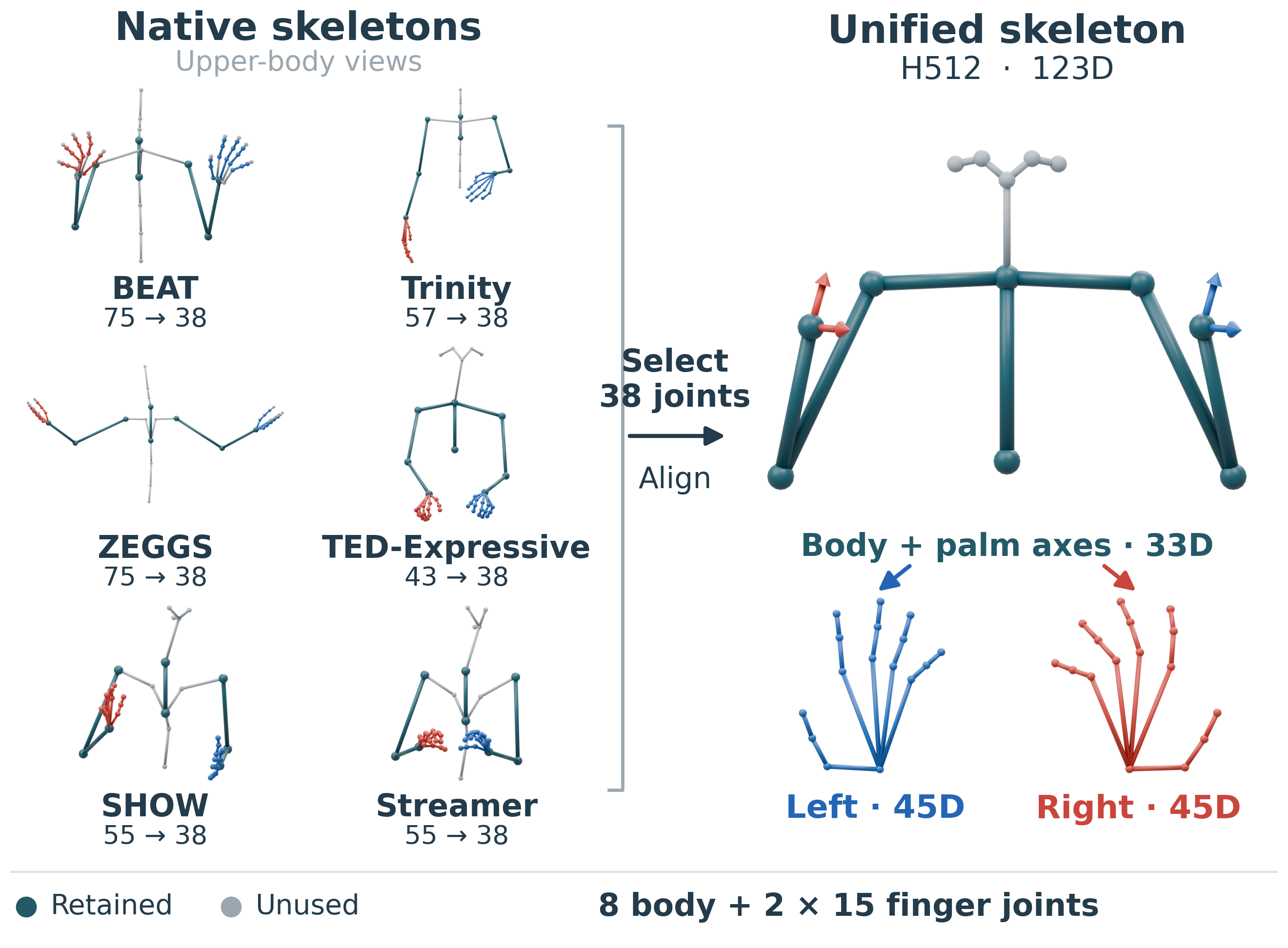}
\caption{
Unified motion representation across the six source datasets.
Motion is first evaluated on each native kinematic hierarchy, and
38 anatomically corresponding joints are then selected and aligned
to a common skeleton. The unified
representation contains 33 dimensions for the upper body and palm
axes and 45 dimensions for each hand, resulting in 123 dimensions.
}
\label{fig:unified_skeleton}
\end{figure}
The six datasets use different skeleton topologies, body proportions, and rotation conventions, making their native representations incompatible. We therefore manually evaluate each motion on its original kinematic hierarchy and extract anatomically corresponding Cartesian landmarks, then align them to a common upper-body and hand skeleton. We perform forward kinematics before simplification, so intermediate joints still contribute to the retained geometry.

To remove dataset-specific bone lengths and rotation conventions, we represent motion using 3D bone directions. For generation, seven torso and arm directions describe the upper body, each palm is represented by its forward and normal axes, and finger directions are expressed in the corresponding palm frame. This results in a 123-dimensional motion representation at 15~Hz as shown in Fig. \ref{fig:unified_skeleton}. The explicit palm axes resolve the axial orientation that cannot be recovered from a single bone direction.

We further verify that this representation preserves the geometry required for robot retargeting. Reconstruction introduces less than $0.6^\circ$ average palm-orientation error on BEAT, Trinity, and ZEGGS, while wrist positions remain effectively unchanged. After retargeting to the G1, GR3, and Reachy2 humanoid robots, the mean differences in shoulder, elbow, and wrist joint trajectories remain below $1^\circ$ across all evaluated source-robot combinations.

\subsection{Spatial Representation and Dataset Construction}

The source datasets contain motion but no corresponding scene geometry or free-space annotations. We therefore construct spatial supervision directly from wrist trajectories, which compactly capture where a gesture is expressed and how far it reaches. As task-level endpoints, the wrists provide direct control over gesture extent while leaving some flexibility to the elbow and shoulder joints. This is desirable for a redundant humanoid arm, where similar wrist motion can often be realized by multiple proximal-arm configurations.

The feasible regions of these proximal links are generally different from the wrist workspace. For example, a wrist may extend beyond a table edge while the shoulder remains behind it, and the elbow may pass below the tabletop outside its boundary. Applying the same box to the entire upper body can impose unnecessary or unnatural constraints. A whole-upper-body envelope is also less informative about hand placement because its boundaries are often dominated by relatively stationary proximal joints: in our cross-recording analysis, nearest upper-body envelopes differ by only 1.19~cm at the median, compared with 11.59~cm for the corresponding wrist envelopes.

We treat the wrist box as a task-level spatial condition, not a complete collision-free region. It only constrains wrist placement so that during applications, the safe region of the movement can be determined by the upstream policy. The generator still predicts upper-body motion; an embodiment-specific quadratic program (QP) retargets it under soft workspace and configured hard physical constraints.

For each motion segment, we compute the smallest torso-centered, axis-aligned box containing both wrist trajectories. The six boundaries are normalized by a canonical arm length of $69$~cm. The box specifies gesture extent without determining which hand moves, its trajectory, or the corresponding arm and finger configuration.

A practical issue arises when envelopes are computed from fixed temporal windows: nearly identical overlapping windows can receive different labels when an outward wrist movement falls just outside one window. To reduce this effect, each training example contains four observed frames and 30 target frames, with four additional future frames used only to test envelope stability. We discard windows whose envelope changes sharply after this extension, together with segments dominated by stationary wrists or silence. These additional frames are never provided to the generator.

Starting from 799,996 candidate windows, this procedure retains 378,438 examples. We split the data by complete recording to prevent overlapping windows from appearing in both training and test sets, resulting in 359,088 training and 19,350 held-out examples.

\section{Method}

\begin{figure*}[h]
  \centering
  \includegraphics[width=\linewidth]
  {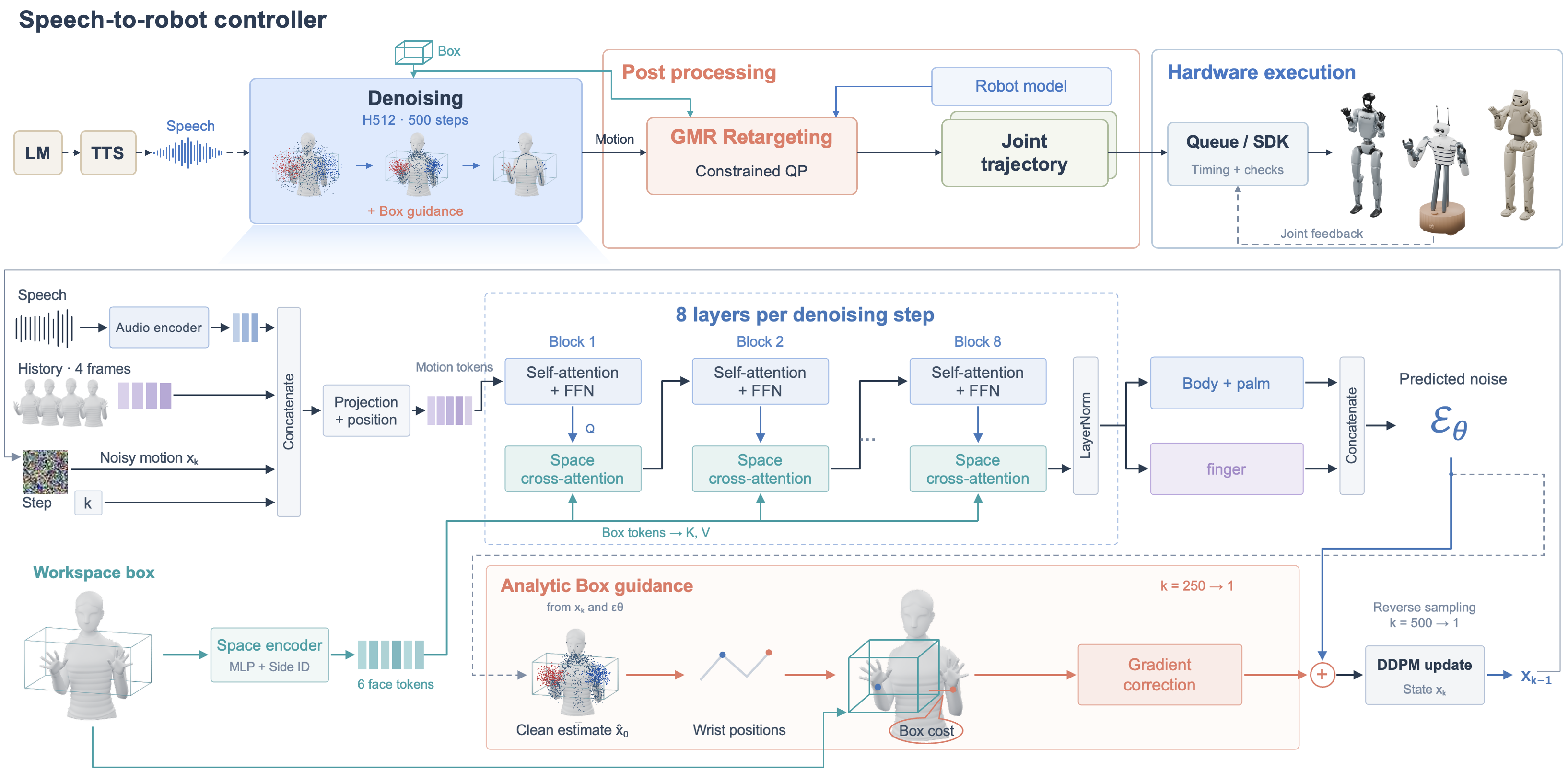}
  \caption{
  Overview of the speech-to-robot framework.
  Speech, motion history, and a workspace box condition a diffusion
  model to generate future gesture motion. The workspace is encoded as six
  spatial tokens and injected through cross-attention, while analytic
  guidance during denoising further corrects wrist motion toward the
  requested workspace. The generated motion is then retargeted to the target
  robot and executed under embodiment-specific constraints.
  }
  \label{mainfigure}
\end{figure*}

Our framework generates co-speech gestures that adapt to a specified workspace and can be executed on different robot embodiments. As illustrated in Fig.~\ref{mainfigure}, given speech, recent motion history, and a workspace box, GestAdapt generates motion with a workspace-conditioned diffusion model, applies analytic denoising guidance, and retargets the result under embodiment-specific execution constraints.

\subsection{Problem Formulation}
\label{sec:problem}
Given speech audio $A$, recent motion history $H$, and a workspace $W$, we aim to generate a future gesture sequence $X$ as
\begin{equation}
X \sim p_\theta(X\mid A,H,W),
\qquad
W=(\ell_x,u_x,\ell_y,u_y,\ell_z,u_z).
\label{eq}
\end{equation}
Motion is represented at 15~Hz, with $h=4$ history frames and $T=30$
window frames.

$W$ denotes the normalized workspace box specifying the desired region for wrist motion. At inference time, it can designate a selected box for gesturing; selecting $W$ from scene or task information is \textbf{outside} our scope. The workspace constrains only the gesture's spatial extent, without prescribing hand assignment, trajectory, or gesture type, leaving the generator free to produce multiple plausible gestures.

\subsection{Workspace-Conditioned Diffusion Transformer Model}
\label{sec:diffusion_model}

Because co-speech gesture is one-to-many, we use a diffusion Transformer to generate multiple plausible motions conditioned on speech and workspace, as stochastic denoising naturally preserves multiple plausible motion realizations.

At each denoising step, temporal motion tokens combine the current noisy future motion, four observed history frames, frame-aligned speech features extracted from 16\,kHz audio by a jointly trained four-layer 1D convolutional encoder, and the diffusion timestep. The audio encoder produces a 32-dimensional feature for each of the 34 motion frames for temporal alignment. The denoiser contains eight Transformer blocks with hidden dimension $D=512$.

The six boundaries of the workspace are embedded by a shared MLP and combined with a learned boundary identity embedding. After temporal self-attention, each Transformer block injects these workspace tokens through residual cross-attention:
\begin{equation}
Z^{(l)}
=
\widetilde Z^{(l)}
+
\operatorname{CrossAttn}_l
\big(
\operatorname{LN}(\widetilde Z^{(l)}),
C,
C
\big),
\label{eq}
\end{equation}
where $\widetilde{Z}^{(l)}$ contains the temporal motion features and $C \in \mathbb{R}^{6 \times D}$ contains the six workspace tokens. The normalized motion features provide the queries, while $C$ provides the keys and values through separate learned projections. This allows each temporal position to attend to the workspace boundaries throughout denoising.

Holistic co-speech motion contains skeleton components with substantially different motion statistics and relationships to speech. Accordingly, recent holistic gesture models commonly use part-wise representations or decoders rather than modeling all motion variables identically~\cite{yi2023talkshow,liu2024emage}. We adopt a similar part-aware design for a task-specific reason. In our unified representation, the body and palm variables form a compact 33-dimensional component that determines the arm configuration and wrist/palm geometry, where workspace adaptation is primarily expressed, whereas finger articulation occupies the remaining 90 dimensions. We therefore use separate output heads for body/palm motion and finger motion. The Transformer backbone remains shared, preserving temporal and cross-part interactions, while the output projections specialize to the different structures of spatially relevant upper-body motion and high-dimensional finger articulation.

Both branches are trained jointly using the standard dimension-normalized DDPM noise-prediction objective over 500 diffusion steps, with a linear noise schedule from $\beta_1=10^{-4}$ to $\beta_{500}=0.02$. The model is trained for 780,000 updates, and inference uses the full 500-step ancestral DDPM sampler. Our representation ablation in Sec.~\ref{sec:ablation} shows improved motion quality and workspace compliance over the 126D holistic design.

\subsection{Analytic Workspace Guidance}
\label{sec:workspace_guidance}

Workspace conditioning substantially reduces spatial violations, but residual wrist boundary crossings can still occur during sampling. We therefore apply an analytic guidance term at sampling time, following the general idea of test-time guidance in diffusion-based motion generation~\cite{liao2026beyondmimic}. The guidance introduces no additional training objective.

At reverse step $k$, we reconstruct an estimate $\widehat{X}_0$ of the clean motion from the current noisy state $X_k$ and the detached denoiser prediction. Wrist positions are then obtained differentiably through forward kinematics. For each future frame $t$, wrist $s$, and spatial axis $a\in\{x,y,z\}$, we define the workspace violation as
\[
v_{t,s,a}
=
\max\left(
\ell_a-\widehat{w}_{t,s,a},
\widehat{w}_{t,s,a}-u_a,
0
\right),
\]
where $\ell_a$ and $u_a$ denote the lower and upper workspace boundaries. We use the guidance cost
\begin{equation}
\begin{aligned}
C_W
&=
\frac{1}{6T}\sum_{t,s,a}v_{t,s,a}^{2}
+
\left(\max_{t,s,a}v_{t,s,a}\right)^2,\\
\widetilde{\epsilon}_k
&=
\epsilon_k+\lambda_k\nabla_{X_k}C_W ,
\end{aligned}
\label{eq:workspace_guidance}
\end{equation}
For the 500-step reverse process, we set
$\lambda_k=G\sqrt{1-\bar{\alpha}_k}$ for $k\leq250$ and zero otherwise,
with $G=96$ by default. The denoiser prediction is detached when
differentiating $C_W$ through the clean-motion reconstruction and forward
kinematics. We apply no gradient normalization, clipping, or state
projection. Thus, $G=0$ retains workspace conditioning without analytic
guidance, which reduces residual violations but provides no hard feasibility
guarantee.
\begin{table*}[t]
    \centering
    \caption{Quantitative results over 5 runs.}
    \label{tab:main_results}
    \small
    \setlength{\tabcolsep}{4pt}
    \begin{tabular}{lcccc}
        \toprule
        Method
        & $\mathrm{FGD}\downarrow$
        & $\mathrm{BC}\uparrow$
        & Jerk $\downarrow$
        & Violation $\downarrow$ \\
        \midrule
        Trimodal (deterministic)~\cite{yoon2020trimodal}
        & $11.045 \pm 0.000$
        & $0.5283 \pm 0.0000$
        & $10.202 \pm 0.000$
        & $6.064 \pm 0.000$ \\

        StyleGestures\cite{alexanderson2020style}
        & $11.579 \pm 0.274$
        & $\mathbf{0.6356 \pm 0.0019}$
        & $6.526 \pm 0.005$
        & $19.667 \pm 0.189$ \\

        DiffSHEG\cite{chen2024diffsheg}
        & $15.579 \pm 0.957$
        & $0.4889 \pm 0.0038$
        & $5.163 \pm 0.117$
        & $16.481 \pm 0.366$ \\

        Ours w/o Workspace
        & $6.933 \pm 0.419$
        & $0.4989 \pm 0.0076$
        & $4.695 \pm 0.105$
        & $9.815 \pm 0.262$ \\

        \textbf{Ours}
        & $\mathbf{5.019 \pm 0.223}$
        & $0.4515 \pm 0.0036$
        & $\mathbf{4.295 \pm 0.104}$
        & $\mathbf{1.095 \pm 0.017}$ \\
        \bottomrule
    \end{tabular}
\end{table*}
\subsection{GMR-Based Robot Retargeting and Constrained Execution}
\label{sec}

The diffusion model generates motion in a unified body representation rather than robot joint space. Here, we use GMR\cite{joao2025gmr} to map each generated sequence to the target embodiment. Generated upper-arm and forearm directions and palm orientations are transformed into the robot frame using embodiment-specific coordinate calibration and link lengths. GMR then solves for robot joint configurations using the robot kinematic model, Jacobians, and joint limits.

The requested workspace is registered once in the robot base frame and converted from the normalized generation coordinates to a metric box using the embodiment-specific scale. At each frame, we augment the existing GMR/Mink tracking problem with soft workspace constraints:
\begin{equation}
\begin{aligned}
    \min_{\Delta q, \, s \geq 0} \quad & \frac{1}{2}\Delta q^{\mathsf{T}} H_{\mathrm{IK}} \Delta q + c_{\mathrm{IK}}^{\mathsf{T}} \Delta q + \frac{1}{2} s^{\mathsf{T}} \Lambda s \\
    \mathrm{s.t.} \quad & A_W \Delta q - s \leq b_W, \\
    & A_{\mathrm{hard}} \Delta q \leq b_{\mathrm{hard}},
\end{aligned}
\label{eq:workspace_qp}
\end{equation}
where $\Delta q$ is the joint position increment and $s \geq 0$ represents the slack variables introduced to soften the workspace constraints. The matrix $H_{\mathrm{IK}}$ and vector $c_{\mathrm{IK}}$ define the quadratic and linear costs of the baseline inverse kinematics (IK) objective. The diagonal weight matrix $\Lambda$ penalizes workspace slack, trading off tracking accuracy and workspace adherence. The matrices $A_W, A_{\mathrm{hard}}$ and vectors $b_W, b_{\mathrm{hard}}$ denote the Jacobian-based linear mappings and upper bounds for the soft workspace constraints and the hard physical constraints (e.g., joint limits and self-collisions), respectively. Workspace constraints are soft, whereas joint ranges, per-frame motion limits, and configured self-collision constraints remain hard.

Generation and execution therefore solve different parts of the problem. The diffusion model can change the gesture itself in response to the target space, while GMR resolves that motion for a particular robot and maintains embodiment-specific feasibility. Accordingly, we evaluate generation-stage wrist containment separately from robot-stage tracking and execution feasibility.

\section{Experiments}

\subsection{Metrics}

Our primary goal is that, when the workspace changes, the generated motion should remain plausible under the distribution of real human gestures, remain responsive to the accompanying speech, and stay smooth, while placing the wrists within the prescribed spatial region. Therefore, we evaluate these complementary properties using four metrics.

\textbf{FGD} (Fréchet Gesture Distance) measures the distributional discrepancy between generated and real gestures in the feature space of a pretrained gesture encoder. We compute FGD between the generated motions and the corresponding real motions in the held-out test set; a lower FGD value indicates that the generated samples remain closer to the unseen real gesture distribution. \textbf{BC} (Beat Consistency) measures speech-motion synchronization by comparing the temporal locations of motion beats with acoustic beats extracted from the corresponding audio. \textbf{Jerk} measures the temporal smoothness of the generated wrist trajectories through their third-order temporal derivative. Finally, \textbf{Violation} measures workspace compliance by computing the maximum distance by which either wrist moves outside its prescribed workspace, averaged over the test set.

\subsection{Quantitative Results}
We compare our full model against three representative external baselines that can be retrained under a common experimental protocol without altering their core formulations. Specifically, Trimodal~\cite{yoon2020trimodal}, StyleGestures~\cite{alexanderson2020style}, and DiffSHEG~\cite{chen2024diffsheg} are retrained on the same six datasets and unified motion representation. We restrict the comparison to methods that can be adapted to this setting with minimal structural changes; methods tightly coupled to dataset-specific motion representations or auxiliary inputs would require substantial modification, making a controlled comparison difficult to interpret. We additionally evaluate \textit{Ours w/o Workspace}, an internal control that uses the same gesture-generation backbone and training setup as the full model but excludes workspace-conditioned design and analytic workspace guidance. This provides a more direct comparison for isolating the effect of workspace-conditioned generation from the underlying motion-generation architecture.

As shown in Table~\ref{tab:main_results}, ours w/o Workspace achieves an FGD of 6.933. The full workspace-aware generation pipeline reduces FGD to 5.019 and mean maximum workspace violation from 9.815 cm to 1.095 cm. Table \ref{tab:main_ablation} further distinguishes learned conditioning from analytic guidance: without guidance, the workspace-conditioned model already achieves an FGD of 5.021 and a violation of 1.272 cm, while the default guidance further reduces violation to 1.095 cm.

Although our model has a lower BC, we report BC for comparability with prior work and interpret it only as a diagnostic of rhythmic alignment rather than perceptual quality. Prior studies have shown limited agreement between objective gesture metrics and human judgments, while conventional BC can be sensitive to spurious motion beats; speech–gesture timing itself also exhibits cross-linguistic variation\cite{yoon2022genea}. We therefore complement it with human evaluation of overall gesture quality, including naturalness, expressiveness, and appropriateness to the accompanying speech.

The comparison with the three external baselines should not be interpreted as fully information-matched. In Table~\ref{tab:main_results}, GestAdapt is conditioned on each test sequence's reference workspace, whereas the external baselines retain their original input modalities. We therefore use these comparisons to assess conditional motion quality and workspace containment rather than to establish superiority under identical input information. In contrast, \textit{Ours w/o Workspace} shares the same underlying generator, training data, and motion representation as the full model, and serves as the controlled comparison for evaluating the effect of workspace-conditioned generation. Sec.~\ref{secuserstudy} further evaluates this distinction under changed workspace conditions; in the robot study, all motion sources are processed through the same workspace-constrained execution pipeline, enabling comparison under identical downstream constraints.

\subsection{User Study}
\label{secuserstudy}
Objective metrics alone do not fully capture perceived gesture quality, as also observed in the GENEA Challenge, where most objective metrics correlated poorly with human judgments~\cite{yoon2022genea}. We therefore conduct two user studies with 30 participants on 12 randomly selected 8-second test clips. In both studies, participants evaluate overall gesture quality considering naturalness, expressiveness, and appropriateness to the accompanying speech. The first study evaluates generated human-skeleton motions under workspace constraints against general motion, while the second evaluates the motions after retargeting and constrained execution on a physical robot. All comparisons are anonymized. Table~\ref{tab:user_study_filtered} reports means and 95\% confidence intervals across participants. We use Ours w/o Workspace as the primary generated-motion control, which allows the study to isolate whether conditioning gesture generation on the target workspace improves perceived motion quality under changed spatial constraints. To isolate the advantage of workspace guidance, Ours w/o Workspace is denoised with the same analytic workspace guidance, followed by the same robot retargeting, which differs from the setting in Table \ref{tab:main_results}.

\textbf{Unseen-workspace generation.}
Each 8-second clip contains four generation windows. We randomly select one window and use its workspace constraint for all four windows, such that three of the four windows must be generated under a workspace different from their original one. The full sequence is generated recursively using the model's previous predictions as motion history, without access to future reference motion. Participants independently rate the ground truth, Ours w/o Workspace, and our generated motions from 1 to 5 for overall gesture quality. Only our full model receives the modified workspace through learned conditioning; the no-conditioning variant receives the same workspace only through analytic sampling guidance. Despite this additional constraint, our method achieves a mean score of $3.24/5$, compared with $2.43/5$ for Ours w/o Workspace and $3.68/5$ for the ground truth, showing that it can substantially alter the gesture to satisfy a new workspace while preserving perceived gesture quality.

\textbf{Robot evaluation.}
We next evaluate whether the generated motions retain their expressiveness under shared workspace-constrained robot retargeting. For each 8-second clip, we randomly sample workspace constraints from unrelated test-set windows. Our method generates motions directly under these constraints, whereas Ours w/o Workspace uses it only through analytic guidance. We then post-process all three using the same workspace constraints, QP post-processing, and robot controller. Participants rank the anonymized robot motions using the same overall-quality criteria. Our method achieves a mean rank of $1.37$ and is ranked first in $69.7\%$ of the trials, compared with $2.11$ and $19.7\%$ for Ours w/o Workspace, and $2.53$ and $10.6\%$ for the ground truth. This indicates that incorporating the target workspace directly into the learned generator improves perceived gesture quality beyond analytic workspace guidance and downstream constrained retargeting alone.

 \begin{table}[t]
      \centering
      \caption{User-study results with 95\% confidence intervals}
      \label{tab:user_study_filtered}
      \small
      \setlength{\tabcolsep}{4pt}
      \begin{tabular}{lccc}
          \toprule
          & Study 1
          & \multicolumn{2}{c}{Study 2} \\
          \cmidrule(lr){2-2}
          \cmidrule(lr){3-4}
          Method
          & Mean score $\uparrow$
          & Mean rank $\downarrow$
          & Top-1 (\%) $\uparrow$ \\
          \midrule
          Ground truth
          & $3.68 \pm 0.20$
          & $2.53 \pm 0.06$
          & $10.6 \pm 3.9$ \\
          Ours w/o Workspace
          & $2.43 \pm 0.18$
          & $2.11 \pm 0.06$
          & $19.7 \pm 4.8$ \\
          \textbf{Ours}
          & $\mathbf{3.24 \pm 0.20}$
          & $\mathbf{1.37 \pm 0.09}$
          & $\mathbf{69.7 \pm 7.1}$ \\
          \bottomrule
      \end{tabular}
  \end{table}

\begin{table*}[h]
    \centering
    \caption{Ablation results over 5 runs.}
    \label{tab:main_ablation}
    \small
    \setlength{\tabcolsep}{4pt}
    \begin{tabular}{lccc}
        \toprule
        Model or setting
        & FGD $\downarrow$
        & Jerk $\downarrow$
        & Violation $\downarrow$ \\
        \midrule

        \multicolumn{4}{l}{
            \textit{(a) Six-corpus model versus single-corpus experts}
        } \\
        \textbf{Ours}
        & $\mathbf{5.019 \pm 0.223}$
        & $4.295 \pm 0.104$
        & $\mathbf{1.095 \pm 0.017}$ \\
        BEAT expert
        & $11.211 \pm 0.241$
        & $6.707 \pm 0.044$
        & $2.950 \pm 0.040$ \\
        Trinity expert
        & $11.918 \pm 0.446$
        & $5.790 \pm 0.086$
        & $5.142 \pm 0.098$ \\
        SHOW expert
        & $12.453 \pm 0.501$
        & $4.935 \pm 0.078$
        & $3.472 \pm 0.106$ \\
        ZEGGS expert
        & $12.862 \pm 0.405$
        & $7.671 \pm 0.114$
        & $7.507 \pm 0.120$ \\
        Streamer expert
        & $13.979 \pm 0.804$
        & $\mathbf{3.286 \pm 0.035}$
        & $2.176 \pm 0.036$ \\
        TED expert
        & $14.230 \pm 0.487$
        & $15.073 \pm 0.387$
        & $2.651 \pm 0.082$ \\

        \midrule
        \multicolumn{4}{l}{
            \textit{(b) Analytic denoising-guidance strength}
        } \\
        $G=0$
        & $5.021 \pm 0.221$
        & $\mathbf{4.291 \pm 0.101}$
        & $1.272 \pm 0.015$ \\
        $G=64$
        & $5.020 \pm 0.222$
        & $4.294 \pm 0.104$
        & $1.140 \pm 0.016$ \\
        $\mathbf{G=96}$ (Ours)
        & $\mathbf{5.019 \pm 0.223}$
        & $4.295 \pm 0.104$
        & $\mathbf{1.095 \pm 0.017}$ \\
        $G=128$
        & Diverged
        & Diverged
        & Diverged \\

        \midrule
        \multicolumn{4}{l}{
            \textit{(c) Motion representation}
        } \\
        Holistic representation
        & $5.942 \pm 0.389$
        & $4.602 \pm 0.100$
        & $1.679 \pm 0.038$ \\
        \textbf{Factorized representation (Ours)}
        & $\mathbf{5.019 \pm 0.223}$
        & $\mathbf{4.295 \pm 0.104}$
        & $\mathbf{1.095 \pm 0.017}$ \\

        \bottomrule
    \end{tabular}
\end{table*}

\subsection{Ablation Studies}
\label{sec:ablation}
We ablate three design choices: multi-corpus training, analytic denoising guidance, and the motion representation. The results are presented in Table~\ref{tab:main_ablation}.

\textbf{Multi-corpus training.}

A single corpus covers only a limited range of gesture patterns and spatial extents. Since our workspace labels are derived from observed motion, each corpus also provides limited workspace-motion combinations. All corpus-specific experts are trained to convergence and evaluated on the same pooled test set and target workspaces. Although the six-corpus model also achieves lower FGD, the more important difference is workspace compliance: its mean maximum violation is $1.095$ cm, compared with $2.176$--$7.507$ cm for the single-corpus experts. This suggests that, under unfamiliar target boxes, models trained on limited spatial support tend to preserve corpus-specific motion patterns at the expense of workspace compliance, motivating broader cross-corpus training.

\textbf{Analytic denoising guidance.}
With learned workspace conditioning alone (\(G=0\)), the mean maximum violation is already 1.272 cm. Analytic guidance further reduces it to 1.095 cm at \(G=96\), with negligible changes in FGD and jerk, indicating that workspace adaptation is primarily learned by the generator while guidance mainly corrects residual violations; \(G=128\) destabilizes sampling.

\textbf{Motion representation.}
 We compare a 126D holistic representation using world-frame directions for all 42 bones with our 123D design, which explicitly represents palm axes and expresses finger directions in the corresponding palm frame, with separate body/palm and finger output heads. The factorized design improves FGD from 5.942 to 5.019 and violation from 1.679 to 1.095 cm.

\section{CONCLUSION}

We presented GestAdapt, a workspace-conditioned framework for robot co-speech gesture generation. By unifying six co-speech datasets with wrist-envelope supervision, GestAdapt generates gestures conditioned on speech, recent motion, and target region. Analytic guidance reduces wrist violations, while constrained retargeting handles embodiment-specific gesture execution requirements. Quantitative evaluation shows low workspace violations and favorable motion performance. In the user study, GestAdapt outperforms its no-workspace counterpart and remains close to the reference motions. Participants prefer GestAdapt to both alternatives after identical workspace-constrained robot retargeting and execution. These findings support incorporating the target workspace into gesture generation rather than introducing it only through downstream correction. Current limitations include the generation of wrist boxes and the lack of whole-body collision guarantees; future work will extend the method toward scene-level geometry and whole-body collision constraints.

\section*{ACKNOWLEDGMENT}

This work was supported by the Dutch Research Council (NWO) through an NWO XS grant (Grant ID FLRUI47134).

\bibliographystyle{IEEEtran}
\bibliography{ref}

\end{document}